# Victory Sign Biometric for Terrorists Identification

Ahmad B. A. Hassanat[a]∗, Mahmoud B. Alhasanat [b], Mohammad Ali Abbadi[a], Eman Btoush [a], Mouhammd Al-Awadi[a], and Ahmad S. Tarawneh[a]

[a]*IT Department, Mu'tah University, Mu'tah – Karak, Jordan*
[b]*Department of Civil Engineering, Al-Hussein Bin Talal University, Maan, Jordan*

ABSTRACT

Covering the face and all body parts, sometimes the only evidence to identify a person is their hand geometry, and not the whole hand- only two fingers (the index and the middle fingers) while showing the victory sign, as seen in many terrorists videos. This paper investigates for the first time a new way to identify persons, particularly (terrorists) from their victory sign. We have created a new database in this regard using a mobile phone camera, imaging the victory signs of 50 different persons over two sessions. Simple measurements for the fingers, in addition to the Hu Moments for the areas of the fingers were used to extract the geometric features of the shown part of the hand shown after segmentation. The experimental results using the KNN classifier were encouraging for most of the recorded persons; with about 40% to 93% total identification accuracy, depending on the features, distance metric and K used.

Keywords— Hand geometry; Hand Segmentation; terrorist identification; victory sign; hand shape biometric

∗ Corresponding author. Tel.: +962-798-897-192; fax: +962-323-75540; e-mail: hasanat@mutah.edu.jo, ahmad.hassanat@gmail.com

## 1. Introduction

Recently the world has witnessed a number of terrorist attacks. Such inhumane actions are applauded by terrorists from different races and religions, those who appear in videos covering all their body parts, changing their voices and holding up victory signs, by raising the index and the middle fingers.

Hand shape biometrics is the ensemble of methods and techniques used for identifying a person depending on the hand silhouette and/or geometric features, which normally include finger widths, lengths, angles, ratios, etc). A typical hand shape recognition system uses a camera or scanner to acquire the hand image of a person, which, with some features extraction methods, is compared with the templates stored in a database to identify a person (Duta 2009).

This paper investigates to what extent a computer system can identify a person (e.g. a terrorist) from only two fingers (their victory sign), because this image might be the only identifying evidence they provide. Perhaps the geometric features are also the only features that can be used in such a case, because the hand is far away from the camera, and therefore the camera will not be able to capture the fine minutiae of the hand and fingers.

Biometric based on the whole hand geometry is a 'hot' research area (Amayeh, Bebis and Hussain 2010), (Pavešić, Ribarić and Ribarić 2004), (Sanchez-Reillo, Sanchez-Avila and Gonzalez-Marcos 2000). However, identifying a person using a small part of the hand (such as the victory sign) is a challenging task, and has, to the best of our knowledge, never been investigated in the literature.

For the purpose of this paper, we have created a new victory sign hand image database (VSHI), consisting of hand images of 50 different users (male and female). This database was used in our investigation, particularly for training and testing.

The rest of this paper presents some of the previous work related to hand shape biometrics, describes the database used for this investigation, presents the methods used for segmentation and features extraction, presents and discusses the results, and finalises the conclusions along with a description of proposed future work related to this study.

## 2. Literature review

Hand shape biometrics is attractive to researchers for the following reasons (Duta 2009):

1) Capturing Hand shape is relatively user friendly (non-intrusive), using low-cost sensors such as mobile camera (Amayeh, Bebis and Erol, et al. 2006), (Kumar, et al. 2006).

2) High resolution images are not required to get the geometry features from the hand shape, and the size of template is relatively small (Sidlauskas 1994).

3) Unlike fingerprints, for instance, this type of biometric has no criminal connotation, and therefore is more acceptable to the users (Kukula and Elliott 2006)

4- Using high quality images, more distinctive features such as palm and fingers minutiae can be integrated to the system (Dutagac, Sankur and Yoruk 2008).

5- Low computational algorithms (Sanchez-Reillo, Sanchez-Avila and Gonzalez-Marcos 2000).

Different methods have been introduced in the literature for the hand shape biometric, and each has its own cons and pros according to speed, accuracy, cost, users acceptance, etc. These include the work of (Sanchez-Reillo, Sanchez-Avila and Gonzalez-Marcos 2000), who introduced a set of geometric features, which can be divided into four different categories: Widths, Heights, Deviations and Angles. Their method achieved 97% success, using Gaussian Mixture Modeling on a database of 20 persons with 10 hand images for each.

Ma and co-workers (Ma, Pollick and Hewitt 2004) proved that B-Spline curves can accurately record the shape of fingers. Varchol and Levicky (Varchol and Levicky 2007) employed 21 geometric features for fingers widths and heights in addition to palm measurements. Using different metrics and Gaussian Mixture Modeling, their method achieved about 85% accuracy.

Aragonès used three approaches for features extraction: holistic, geometric and feature-based, the extracted features were 256 for the palm and 121 for fingers, and the accuracy achieved was in the range of 74% to 98% depending on the method used (Aragonès 2013).

Luque-Baena and co-workers (Luque-Baena, et al. 2013) used Genetic algorithm, Mutual Information and Linear Discriminant Analysis to drastically reduce 400 features of hand to about 50. 97% - 100% accuracy were achieved using different databases.

Kang and Wu (Kang and Wu 2014) used Fourier descriptors and finger area functions, and classified using Euclidean Distance. Their experiments were conducted on Bosphorus Hand Database (Bogazici-University 2015). The error rate was about 4% .

Other researchers who worked on hand shape biometrics include and not limited to: (Dhole 2012), (Guo, et al. 2012), (Basheer and Robinson 2013) and (Hassanat, Al-Awadi, et al. 2015).

To the best of our knowledge, the only approach found in the literature related to identifying persons from their hands based on one part of the hand, is the type of work that is based on identifying persons from their finger knuckle alone. A recent example of such an approach is the work of (Usha and Ezhilarasan 2015), who proposed a new method based on geometric and texture analyses of the finger knuckle.

However, the Finger knuckle will not work for identifying terrorists seen in a digital image or video, for the same reason that we cannot use their fingerprints, as the fine details for such approaches are not retained in such a situation.

A more related approach is the identification of persons using their behavioral characteristics for hand gestures of hand sign language. Fong and co-workers (Fong, et al. 2013) used simple and efficient image processing algorithms, including intensity profiling, color histogram and dimensionality analysis, coupled with several machine learning algorithms. Their experiments conducted on a database of only 4 people performing four sets of hand gestures for the 26 letters according to the American Sign Language. The identification accuracies were in the range of 68.75% to 93.75% depending on the classifier used.

However, the behavioral characteristics for hand gestures of hand sign language will not work for identifying terrorists seen in a digital image or video performing only one sign (the victory sign). This leaves the physiological characteristics of hand shape, particularly the victory sign shape, as the only hope in such situation. Therefore, this paper is best suited to addressing this problem.

## 3. Data and Methods

*3.1. VSHI Database*

A mobile phone camera (8 mega pixel, 3264 x 2448) was used to capture the right-hand victory sign images of 50 persons (male and female) from different age groups in the range [14, 50] years. The images were taken in two different sessions (3 days period was left to record the second session) with five images for each session. The total number of images is 500. All the images of the hand were upright with some rotation allowed around -45 to 45 degrees. The mobile phone camera was upright when the images were taken, and that is why the hands appear horizontal in images. All hand images were in the foreground against a black background to ease hand segmentation. Having known that the shape is not affected by image resizing, we scaled down all images in VSHI to be (0.125) of the actual size, this was just to increase the speed of our experiments.

VSHI was designed mainly for the purpose of this study, which is to identify persons from their victory signs, particularly, the case of terrorists. However, it can be used for other purposes such as: distinguishing between male and female; and identifying the person's age groups. Such purposes are also related to identifying a terrorist, nonetheless, those purposes will not be investigated in this work.

We followed the following rule for naming the image files: the file name starts with (P) followed by the person's number, then the subject gender (male: M or female: F), which is followed by the age (in years) of the person, followed by the session number (S1 or S2), and finally the image number within each session from 1 to 5. Table (1) depicts the naming system of the image files in VSHI database.

We made VSHI database publically available for download at the Mutah University website: https://www.mutah.edu.jo/biometrix.

**Table 1. Naming system used in VSHI database.**

| #example | Session | person | Gender | Age | #Image | Filename |
|---|---|---|---|---|---|---|
| 1 | S1 | P25 | F | 50 | 3 | P25-F-50-S1 (3) |
| 2 | S1 | p14 | M | 14 | 1 | P14-M-14-S1 (1) |
| 3 | S2 | P35 | M | 35 | 3 | P35-M-35-S2 (3) |
| 4 | S2 | P15 | M | 14 | 1 | P15-M-14-S2 (1) |

*3.2. Methods*

A typical hand shape biometric system consists of a sequence of several major steps, including: Image Acquisition, hand Segmentation, Feature Extraction, Training/Testing and storing the Hand Templates for identification. Fig. (1) shows these steps. In this work we opted for such a system to fulfill our main goal.

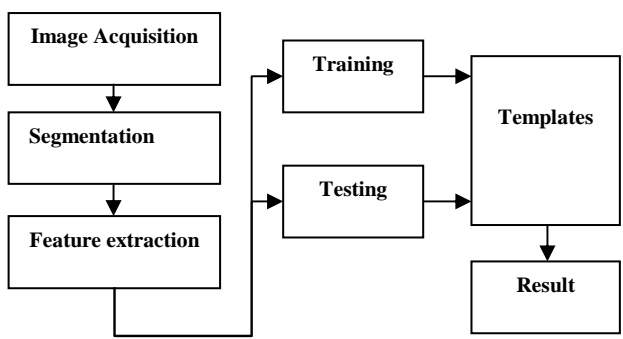

**Fig. 1.** A typical hand shape biometric system

The Image Acquisition step is already done by creating the VSHI database.

Image segmentation is an important processing step in many image, video and computer vision applications. Extensive research has been conducted to find different approaches and algorithms for image segmentation. However, it is still difficult to assess whether one algorithm produces more accurate segmentations than another, whether it be for a particular image or set of images. The most common method for evaluating the effectiveness of a segmentation method is based on subjective evaluation, in which a human visually compares the image segmentation results.

For the purpose of our work, we investigated three approaches to segment the hand, 1) Otsu's method (Otsu 1979), 2) the k-means clustering based on the colour information (RGB color model), and 3) hand segmentation based on colour information using Artificial Neural Network (ANN) (Hassanat, Alkasassbeh, et al. 2015).

The k-means clustering attempts to cluster the image pixels to object/non-object (hand/non-hand) depending on their RGB colour information, using a distance metric, we used two distances, the Euclidean and Hassanat distance (Hassanat 2014).

Otsu's method is based on performing histogram shape-based image thresholding. Both of these segmentation methods assume that the image to be segmented contains two classes of pixels, e.g., foreground (the hand) and background of a different colour. We applied both algorithms on all the images in VSHI; a sample of the segmentation results is shown in Fig. (2).

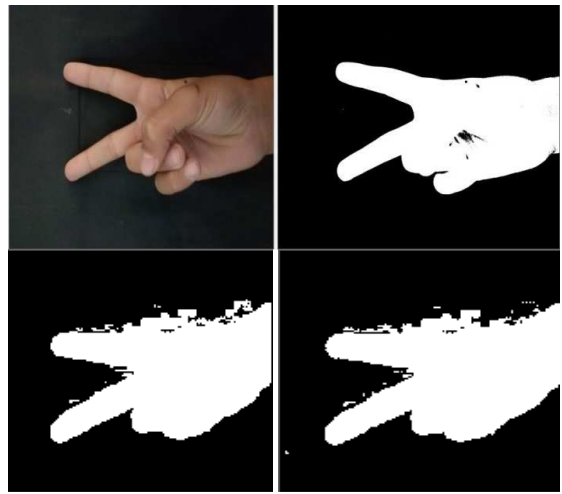

**Fig. 2.** Segmentation results: (up-left) Original; (up-right) Otsu; (down-left) K-means clustering/Euclidean; (down-right) K-means clustering.

As can be seen from Fig. (2) Otsu's method found the optimal segmentation for most of the images in VSHI, while the k-means clustering with both distances was unable to segment any image perfectly. This is due to the similarity (in terms of distance metric) between some pixels in the hand and the background. Otsu was not affected by such similarity because each pixel group (hand and background) belongs to a different area in the image histogram, and even if the difference between any two different pixels is equal to 1, the algorithm was able to segment them correctly because it uses the best threshold. Moreover, the Otsu method employs simple calculations and is much faster than the k-means clustering. However, the Otsu method failed to perfectly segment some images, particularly around some of the contours where there are some pixels affected by the shadow of the hand. Therefore we had to use the third method described in (Hassanat, Alkasassbeh, et al. 2015), which requires the foreground and the background pixels to be trained to object/non-object pixels. By applying this method after training 20 pixels (object/non-object) from 10% of total images in VSHI, we obtained a perfect (100%) segmentation for the hand silhouette. Table (2) shows subjective comparison of various segmentation methods investigated for the purpose of this study.

**Table 2. Comparison of various segmentation methods used on VSHI images.**

| Segmentation method | Results | Pros | Cons |
|---|---|---|---|
| Otsu (Global thresholding) | 96% | Simple, fast, widely used, and stable results. | Affected by some shadow around some fingers. |
| K-mean clustering (local Thresholding) | 0% | Simple and widely used | Bad segmentation, slow (particularly, with large images) |
| (Hassanat, Alkasassbeh, et al. 2015) | 100% | It detects all patterns of VSHI, fast | It needs training, and consumes time in training phase. |

**Features Extraction**

Perhaps the only features that can be targeted (for the goal of our paper) are the geometric features, because the other stronger features might not be available in a digital image of a terrorist. Normally, geometric features are simple to be computed and their usefulness has been proved in biometric problems (Liao and Pawlak 1996).

Limited by the information available (just two fingers and only geometric features), we investigated several methods for extracting distinctive features to be used for identification:

**Method 1 (feature points)**

Five important points were extracted: 1) Two tips of the fingers (the index and middle fingers); 2) One valley (the point between fingers); 3) Two points in the hand palm. Algorithm (1) finds these points described in detail.. The contour of the hand is extracted by dilating the segmented (binary) image and subtracts it from binary image, which results from the segmentation process. Fig. (3) shows this process.

---

*Algorithm 1: finds fingers tips, between fingers point, upper and lower palm points of hand image.*
Input: Binary image
Output 1: five points (x,y)
==========================================
**Step** 1: Read img1 file from a folder of the VSHI database
**Step** 2: img2= segmentation (img1).
**Step** 3: img3=Dilation (img2)-img2.
**Step** 4: (assuming the all hands are horizontals in the images), Scan img3 from top to bottom and left to right, until finding the first white pixel (p), Tip1=p
**Step** 5: Scan img3 from bottom to top and left to right, until finding the first white pixel (p), Tip2=p
**Step** 6: The point between fingers is the one between Tip1 and Tip2, and having the maximum x value (MaxX) by scanning from left to right→BFP
**Step** 7: Scan image from top to bottom at x= MaxX, until finding the first white pixel (p), upper palm point =p→UPP
**Step** 8: Scan the image from bottom to top until finding the first white pixel (p), bottom palm point =p →BPP

---

The Geometric features that can be calculated based on the output points from Algorithm (1) are: 1) Euclidean distance (ED) from Tip1 to the UPP; 2) ED from Tip1 point to BFP; 3) ED from Tip2 to BPP; 4) ED from Tip2 to BFP; 5) ED between UPP and BFP; 6) Triangle area of the middle finger (Equation (1)) and 7) Triangle area of the index finger, as shown in Fig. (3).

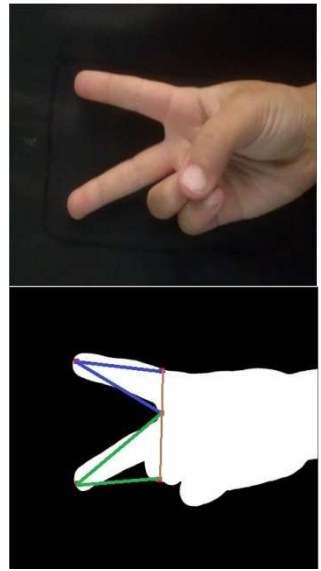

**Fig. 3.** (Left) Original image from VSHI database; (Right) EDs are shown as lines between feature points

Features 6 and 7 can be calculated by

$$Area = \frac{|A_x(B_y-C_y)+B_x(C_y-A_y)+C_x(A_y-B_y)|}{2} \quad (1)$$

where x and y are the coordinates of the points A, B and C of a triangle.

**Method 2 (shape moments)**

The shape moments are used in the literature to describe the shapes of objects because they represent important statistical properties of the segmented object and their efficacy was proved to describe objects (Kang and Wu 2014).

The central $(j,k)^{th}$ moment of a subset of points or pixels (S) is defined by

$$\mu_{jk} = \sum_{(x,y)\in S}^{n}(x-\bar{x})^j(y-\bar{y})^k \quad (2)$$

To make a Moment ($\eta_{jk}$) invariant to both translation and scale, we may divide the corresponding central moment by the $(00)^{th}$ Moment using

$$\eta_{jk} \frac{\mu_{jk}}{\mu_{00}^{(1+\frac{j+k}{2})}} \quad (3)$$

We can now calculate a set of seven invariant Moments, the so-called Hu Moments (Hu 1962). These moments contain information about the geometry of the segmented object, and more importantly, they are invariant to translations, scale changes and rotations. Hu Moments are given by the following

$H1 = \eta 20 + \eta 02$
$H2 = (\eta 20 + \eta 02)^2 + 4\eta^2 11$
$H3 = (\eta 30 + 3\eta 12)^2 + 3(\eta 21 - \eta 03)^2$
$H4 = (\eta 30 + \eta 12)^2 + 3(\eta 21 - \eta 03)^2$
$H5 = (\eta 30 + 3\eta 12)(\eta 30 + 3\eta 12)$
$\quad * [(\eta 30 + \eta 12)^2 - (\eta 21 - \eta 03)^2]$
$H6 = (\eta 20 - \eta 02)[(\eta 30 + \eta 12)^2 - (\eta 21 + \eta 30)^2$
$\quad + 4\eta 11(\eta 30 + \eta 12)(\eta 21 + \eta 03)]$
$H7 = (3\eta 21 - \eta 03)(\eta 30 + \eta 12) * [(\eta 30 + \eta 12)^2$
$\quad - 3(\eta 21 + \eta 03)^2$
$\quad + (\eta 03 - 3\eta 12)((\eta 21 + \eta 03)[3$
$\quad * (\eta 30 + \eta 12)^2 - (\eta 21 + \eta 03)^2]$

In addition to these 7 features, we added another important feature for describing shapes, which is Eccentricity. Eccentricity (E) can be calculated using

$$E = \sqrt{1 - \frac{\lambda_2}{\lambda_1}} \qquad (4)$$

where

$$\lambda_i = \frac{\mu'_{20} + \mu'_{02}}{2} \pm \frac{\sqrt{4\mu'^2_{11} + (\mu'_{20} - \mu'_{02})^2}}{2},$$

and

$$\mu'_{20} = \mu_{20}/\mu_{00}$$
$$\mu'_{02} = \mu_{02}/\mu_{00}$$
$$\mu'_{11} = \mu_{11}/\mu_{00}$$

We calculated the Hu Moments (H1-H7) and (E) for each finger in each image in VSHI, obtaining 16 features to identify each person. Each binary image hand was cut to separate the index finger and the middle finger based on the feature points described in Method (1). Fig. (4) shows the cutting process.

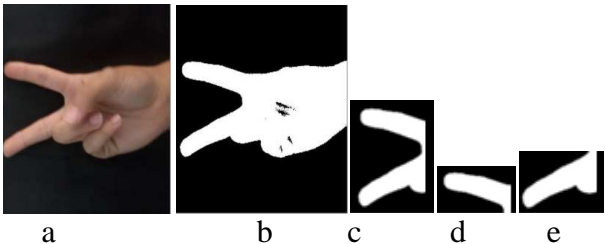

**Fig. 4.** a) Original; b) Binary; c) fingers; d) index finger; e) middle finger

## 4. Results and discussions

To evaluate the identification process for persons in VSHI we conducted 3 experiments, 1) identification of persons using Method (1); 2) identification of persons using Method (2); and 3) identification of persons using concatenated features from Methods (1) and (2).

We worked on Sessions 1 and 2 from the VSHI database, taking 34% of random hands for the testing set, and the remaining 66% were chosen for the training set, for each session. Each experiment was repeated 10 times, with random test and training examples, and the average accuracy was recorded. Session (2) from VSHI was used for validation.

The KNN classifier (with k=1, 3 and 5) was used using Euclidean distance (ED), Manhattan distance (MD) and Hassanat distance (HD). The scale of accuracy of identification are shown in Table (3).

**Table 3. Identifying results using 34% test, and 66% training, results are averaged over 10 runs.**

|  |  | Session 1 | | | Session 2 | | |
|---|---|---|---|---|---|---|---|
| Method | K | ED | MD | HD | ED | MD | HD |
| 1 | 1 | **0.772** | 0.755 | 0.751 | **0.838** | 0.822 | 0.819 |
| 2 | 1 | 0.839 | 0.845 | **0.854** | 0.858 | 0.878 | **0.891** |
| 3 | 1 | 0.915 | **0.925** | 0.924 | 0.893 | 0.917 | **0.920** |
| 1 | 3 | **0.684** | 0.680 | 0.674 | **0.703** | 0.694 | 0.690 |
| 2 | 3 | 0.758 | 0.771 | **0.780** | 0.780 | 0.802 | **0.810** |
| 3 | 3 | **0.773** | 0.769 | 0.771 | 0.798 | 0.817 | **0.822** |
| 1 | 5 | **0.567** | 0.554 | 0.558 | 0.524 | 0.527 | **0.529** |
| 2 | 5 | **0.587** | 0.579 | 0.585 | 0.608 | 0.608 | **0.613** |
| 3 | 5 | **0.664** | 0.654 | 0.652 | 0.609 | 0.621 | **0.624** |

**Table 4. Validation of the system, using session (1) for training, and session (2) as validation dataset; with K=1.**

| Method | ED | MD | HD |
|---|---|---|---|
| 1 | 0.399 | 0.486 | **0.731** |
| 2 | 0.609 | **0.680** | 0.680 |
| 3 | 0.625 | **0.747** | 0.739 |

As can be seen from the results using a simple classifier for 50 different classes, the accuracy is good enough to identify persons, and therefore there is a great potential for this approach to be used for the purpose of identifying terrorists, that is, if the victory sign were the only identifying evidence.

Obviously, the higher accuracy was obtained when we fused features from methods (1) and (2), which we called method (3); hence such a fusion provides more information about the subject.

It can be noticed also that the identification accuracy is decreased while the number of nearest neighbors (K) is increased. This is due the number of examples for each class, having known that each subject has only 5 examples in each session, and those are divided between the testing and training sets, leaving less than 5 examples in the training set. Hence by enforcing the nearest 5 neighbors; unrelated subjects are counted for deification, which increases the error rate.

The results using Hassanat distance were almost the best compared to the other distance, this is due to the ability of this distance metric to deal with noisy data and outliers (Hassanat 2014).

A closer look at the results in Tables (3) and (4) shows that Method (1) does not perform well every time, and this is due to the appearance of the other fingers in the segmentation. As we did not eliminate those fingers, which are bent during performing the victory sign, this produces unexpected and unpredictable measurements of the index or the middle finger. Moreover, the detection process of the fingers from the binary image of the hand (finding the key points using top search to detect the index finger and bottom search to detect the middle finger) is naive and needs to be enhanced, to retain only the required shape of the fingers. For example, this method completely failed to segment the fingers of subject (46) in both sessions (see Fig. (5)) giving unpredictable features for all methods.

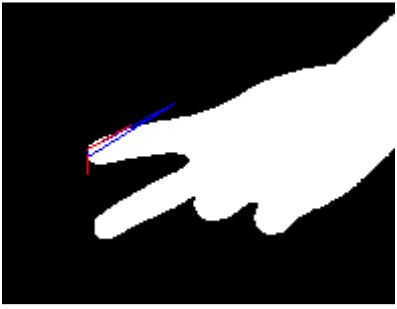

**Fig. 5.** Error example of segmenting the area of the fingers based on the feature points

A closer look at the results in Table (4) shows that the validation results are weaker than that of Table (3), and this is due to the different behavior of the users when they perform their victory sign; hence some of the users produced a different victory sign in the next session, particularly the way they bent their other fingers, and because of the naïve segmentation of the fingers, the signatures from session (2) were more or less different from those of session (1).

There are other limitations of this work those including missing important features from the fingers, such as the widths and the lengths of both fingers. Such information might increase the accuracy of the identification system.

## 5. Conclusion and future work

This work proposes a new approach for identifying persons, particularly terrorist, from their geometric features of victory sign alone, as this sign might be the only information available about such persons. For this purpose we created a new database for the victory sign, and investigated some simple approaches to identify persons from those images. Since we do not have other information like the face of the terrorist to associate the victory sign with, we assume that police or other authorities already having images and hand images for suspects to be associated with the given victory sign of an unknown terrorist.

The proposed method can be used for person's identification in general, however, in general person's identification we have stronger evidences that we can use such as face, the whole hand, fingerprint etc. while in recognizing a terrorist, sometimes the victory sign is the only available evidence.

Although the used methods were simple, our experiments show great potential for identifying persons having weak evidence (the shape of the victory sign). The experiments also show some limitations of this work, which can be summarized by 1) the weak features used and 2) the weak detection of the fingers used. These limitations will be addressed in the future work. The weak features will be supported by other features such as the widths and lengths of the fingers. And the weak segmentation of the fingers will be changed using the regression analysis.